\documentclass[10pt,twocolumn,letterpaper]{article}

\usepackage[pagenumbers]{wacv} 

\usepackage[accsupp]{axessibility}  

\usepackage{graphicx}
\usepackage{amsmath}
\usepackage{amssymb}
\usepackage{booktabs}

\usepackage{comment}
\usepackage{multirow}
\usepackage{amsfonts}
\usepackage{makecell}
\usepackage{booktabs,siunitx}
\usepackage[table,svgnames]{xcolor}

\usepackage[pagebackref,breaklinks,colorlinks]{hyperref}

\usepackage[capitalize]{cleveref}
\crefname{section}{Sec.}{Secs.}
\Crefname{section}{Section}{Sections}
\Crefname{table}{Table}{Tables}
\crefname{table}{Tab.}{Tabs.}

\def\wacvPaperID{244} 
\def\confName{WACV}
\def\confYear{2024}

\begin{document}

\title{Self-Supervised Representation Learning with Cross-Context Learning between Global and Hypercolumn Features}

\author{Zheng Gao \and Chen Feng \and Ioannis Patras \and \\
	Queen Mary University of London, Mile End Road, London, E1 4NS\\
	{\tt\small \{z.gao, chen.feng, i.patras\}@qmul.ac.uk}
}
\maketitle

\begin{abstract}
   Whilst contrastive learning yields powerful representations by matching different augmented views of the same instance, it lacks the ability to capture the similarities between different instances. One popular way to address this limitation is by learning global features (after the global pooling) to capture inter-instance relationships based on knowledge distillation, where the global features of the teacher are used to guide the learning of the global features of the student. Inspired by cross-modality learning, we extend this existing framework that only learns from global features by encouraging the global features and intermediate layer features to learn from each other. This leads to our novel self-supervised framework: \textbf{c}ross-context learning between \textbf{g}lobal and \textbf{h}ypercolumn features (CGH), that enforces the consistency of instance relations between low- and high-level semantics. Specifically, we stack the intermediate feature maps to construct a ``\textit{hypercolumn}'' representation so that we can measure instance relations using two contexts (hypercolumn and global feature) separately, and then use the relations of one context to guide the learning of the other. This cross-context learning allows the model to learn from the differences between the two contexts. The experimental results on linear classification and downstream tasks show that our method outperforms the state-of-the-art methods.
\end{abstract}

\section{Introduction}
Representation learning has become a challenging and active topic in computer vision, capable of learning representations that can be transferred to various downstream tasks, such as classification, object detection, segmentation, etc~\cite{ji2019invariant,khosla2020supervised,cvpr19unsupervised,hjelm2019learning,dwibedi2021little}. Due to the capability of leveraging massive amounts of data without requiring annotations, self-supervised representation learning, in particular, has shown the potential to learn representations that generalize well on various downstream tasks.

Contrastive learning has shown promising results in the self-supervised representation learning~\cite{zhao2020makes,he2020momentum,grill2020bootstrap,xie2021propagate,wang2021dense,feng2022adaptive}. Contrastive learning aims to learn invariant representations for different views of the same image instance (augmented views should have similar features while different instances are forced to have dissimilar features). Therefore, it lacks the ability to capture similar semantics shared between different instances~\cite{zheng2021ressl,guo2022hcsc}. As a result, it suffers from the so-called ``class collision problem''~\cite{chen2022perfectly,li2021prototypical}. Recent methods aim to alleviate this limitation by capturing similarity relationships among instances based on the knowledge distillation framework where the student is trained to predict the target similarity distribution from the teacher~\cite{zheng2021ressl,Gidaris_2021_CVPR}. The similarity relationships are typically measured with the cosine similarities between the input and the samples in the memory bank, which are normalized with a softmax operation. This leads to a probabilistic distribution where similar instances are emphasized so that the student is trained to produce correlated features for similar samples. However, these methods are limited to use the global features (after the global average pooling) of the teacher to guide the learning of the global features of the student. We term this line of works as ``\textit{global-context learning}'' in this paper.

Works in cross-modality learning~\cite{alwassel2020self,peng2022balanced} have shown that the learning paradigm of one modality can benefit from cross-modal information from multiple modalities. While an additional modality is not available when considering visual-only data, we argue that \textbf{whilst features from the intermediate layers and global features from the final layer are correlated, they encode semantics at different levels of abstraction} -- the earlier layers capturing lower-level details while the latter layers capturing higher-level semantics. The differences between intermediate layers and global features can facilitate the learning of both compared with global-context learning in current works. Inspired by this, we treat the intermediate layers and global features as two contexts and propose a cross-context learning strategy where these two contexts learn from each other. More specifically, we construct a ``\textit{hypercolumn}'' representation~\cite{hariharan2015hypercolumns} by stacking the concatenation of intermediate feature maps as the context of the intermediate layers. Then we measure similarity relationships among instances using two contexts (hypercolumn and global feature) separately, and \textbf{use the similarity relationships of one context as supervision to guide the learning of the other}. This leads to a novel self-supervised framework--cross-context learning between global and hypercolumn features (CGH)--that learns representations by capturing cross-context information from global features and hypercolumns.

We highlight our proposed CGH framework degenerates to ReSSL~\cite{zheng2021ressl} as a special case of global-context learning when the hypercolumn only uses the last layer. The linear classification results on ImageNet show that the proposed CGH outperforms MoCo-v2~\cite{chen2020mocov2} and ReSSL~\cite{zheng2021ressl} by $3.0\%$ and $1.2\%$ respectively with 200 epochs pre-training.

The contributions of this paper can be summarized as follows:
\begin{itemize}
	\item We address the class collision problem in contrastive learning by capturing similarity relationships among instances with the knowledge distillation framework. In contrast to previous methods that are limited to global feature-based instance relations, we propose a novel cross-context learning scheme in which two contexts, one constructed from intermediate layers (hypercolumn) and one from global features, are used to supervise each other.
	\item We show that using the proposed hypercolumn-based representations to capture instance relationships is beneficial and leads to learning better global representations. Precision-recall graphs on the similarity distributions show that this leads to significantly higher recall at very similar precision levels (\cref{sec:CGH-effect}).
	\item Our method is simple and effective. The experimental results on self-supervised benchmarks show that our method achieves superior performance on linear evaluation and downstream tasks compared with state-of-the-art methods, which demonstrates the effectiveness of the cross-context learning strategy.
\end{itemize}

\begin{figure*}[htb]
	\centering
	\includegraphics[width=0.9\linewidth]{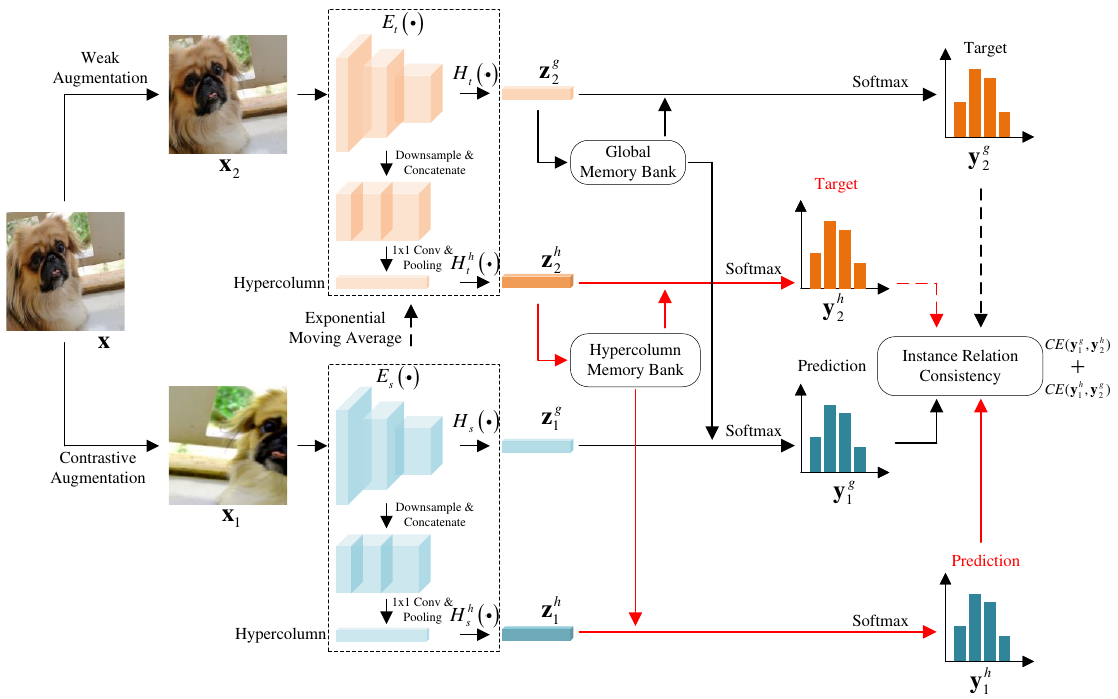}
	\caption{Overview of the proposed CGH framework. We adopt a knowledge distillation framework where the teacher is the exponential moving average of the student. A heavily corrupted view $\mathbf{x}_1$ is fed into the student $E_s$ to obtain both a hypercolumn embedding $\mathbf{z}^h_1$ and a global embedding $\mathbf{z}^g_1$ while a weakly augmented view $\mathbf{x}_2$ is passed to the teacher $E_t$ to obtain a hypercolumn embedding $\mathbf{z}^h_2$ and a global embedding $\mathbf{z}^g_2$. The embeddings are used to measure the similarity relationships between the augmented views $\mathbf{x}_1$, $\mathbf{x}_2$ and the samples in the memory bank -- this leads to a similarity distribution. We enforce two instance relations alignments: ``\textit{global-hypercolumn alignment}'' and ``\textit{hypercolumn-global alignment}'', which are detailed in the text.}
	\label{fig:CGH-overview}
\end{figure*}

\section{Related work}
\subsection{Contrastive learning}
Contrastive learning based on the Siamese structure aims to learn representations by ensuring positive pairs stay close in the latent space and keeping the negative pairs far away~\cite{li2020addressing,he2020momentum,chen2020mocov2,pmlr-v119-chen20j,chen2020big,henaff2020data,robinson2021contrastive,rame2021ixmo}. To achieve this, contrastive learning maximizes the correlation between different transformed versions of the same image in the latent space and minimizes that of negative pairs~\cite{pmlr-v119-chen20j}. MoCo~\cite{he2020momentum} performs contrastive learning through a dictionary lookup, whilst SimCLR~\cite{pmlr-v119-chen20j} simplifies MoCo's sampling strategy by generating negative samples from the current batch instead of maintaining a memory bank. However, it can be a challenge to generate meaningful negative samples. Therefore, non-contrastive methods without negative pairs are developed~\cite{grill2020bootstrap,chen2020exploring} by using techniques like stop-gradient and prediction head to prevent collapsing so that the Siamese network doesn’t produce a constant output.

\subsection{Deep clustering}
Contrastive learning forces every instance to be assigned to a distinct class by pushing different instances apart. By contrast, deep clustering based methods~\cite{li2021prototypical,guo2022hcsc} map similar instances to the same class to solve the class collision problem. Typically, deep clustering based methods leverage clustering algorithms like K-Means~\cite{lloyd1982least} to assign a pseudo label for each instance so that similar instances can be clustered into the same clustering centroid. DeepCluster~\cite{caron2018deep} uses K-means to generate labels for the samples, which are used as pseudo labels to provide supervisory signals for learning representations. SwAV~\cite{caron2020unsupervised} proposes an online clustering algorithm and enforces the consistency of cluster assignments between different views of the same image. The previous methods only establish a single hierarchy of the images, PCL-v2~\cite{li2021prototypical} discovers the multiple semantic hierarchies of the images and performs instance-wise and instance-cluster contrastive learning to solve the class collision problem. HCSC~\cite{guo2022hcsc} extends the work of PCL-v2~\cite{li2021prototypical} by selecting high-quality positive and negative pairs based on the similarity between the samples and the centroids. However, most of these works are based on a strong assumption that the labels must induce an equipartition of the data~\cite{zheng2021ressl}.

\subsection{Inter-sample relations}
Further works try to alleviate class collision by extending the positive sample pair to a set of positive samples~\cite{koohpayegani2021mean,dwibedi2021little,navaneet2022constrained}. NNCLR~\cite{dwibedi2021little} generates an additional positive pair by finding the nearest neighbor of the input image. MSF~\cite{koohpayegani2021mean} compares the input image with several nearest neighbors stored in a memory bank. CMSF~\cite{navaneet2022constrained} further generalizes the idea in MSF by refining the search space of nearest neighbors so that the search space is correlated to the query image yet has sufficient variances. A close line of works to this paper aim to solve the class collision problem by capturing instance relations based on self-distillation~\cite{kim2021self,zhang2019your}. OBoW~\cite{Gidaris_2021_CVPR} trains the student to predict the similarity distribution over the vocabulary generated by the teacher, which is built upon the local views of the feature maps and works as a codebook. Instead of using the quantized feature map to generate the target, ReSSL~\cite{zheng2021ressl} uses different views for the teacher and student based on the weak-contrastive augmentation strategy. The view through weak augmentation is fed into the teacher to provide a reliable target while the other view (via contrastive augmentation) is passed to the student for prediction. The consistency between different views is enforced. These methods are limited to only learn from global features after the global pooling layer. By contrast, we propose to enforce cross-context learning by using the one of the context (say, intermediate layers) as guidance to learn the other (say, global features).

\section{Methodology}
\subsection{Overview}
An overview of the proposed approach is shown in~\cref{fig:CGH-overview}. The core idea  of our scheme is to learn from cross-context information, one context derived from the global features after the global average pooling and one context derived from a hypercolumn that is constructed by the concatenation of intermediate feature maps. To achieve that, we enforce two instance relations alignments: ``\textit{global-hypercolumn alignment}'' and ``\textit{hypercolumn-global alignment}''. The global-hypercolumn alignment aims to use the hypercolumn of the teacher $E_t$ to generate a target similarity distribution to guide the learning of the similarity distribution based on the global feature of the student $E_s$ while hypercolumn-global alignment uses the global feature of the teacher to create a target distribution for guiding the learning of the similarity distribution based on the hypercolumn for the student. The effectiveness of the cross-context learning is analysed in~\cref{sec:CGH-effect}. We provide the training cost analysis in the supplementary material.

\subsection{Cross-context learning}\label{sec:CGH-cross}
Given an image $\mathbf{x}$, we generate a weakly augmented view $\mathbf{x}_2$ through weak augmentation for the teacher and a heavily augmented view $\mathbf{x}_1$ through contrastive augmentation for the student as in~\cite{Gidaris_2021_CVPR,zheng2021ressl}. Compared with the weak augmentation, the contrastive augmentation is more aggressive and generates heavily corrupted views. The student is trained to adapt to the heavy disturbance and noise introduced by the contrastive augmentation to learn robust representations. By contrast, the teacher generates a stable target based on the less aggressive weak augmentation.

We then proceed to generate the contexts of global feature and hypercolumn for the teacher and the student separately, as shown in~\cref{fig:CGH-overview}. First, $\mathbf{x}_2$ is passed to the teacher encoder $E_t$ to produce the ``\textbf{\textit{global feature context}}'' (after the global average pooling) $\mathbf{h}^g_2 = E_t(\mathbf{x}_2)$. Then $\mathbf{h}^g_2$ is transformed by a global projector $H_t$ to produce a low-dimensional global embedding by $\mathbf{z}^g_2 = H_t(\mathbf{h}^g_2)$ as in~\cite{chen2020mocov2,zheng2021ressl}. As for the hypercolumn of the teacher, let $E^l_t(\mathbf{x}_2) \in \mathbb{R}^{{c_l}\times{h_l}\times{w_l}}$ be the intermediate feature maps of the $l$-th convolutional block, $l \in \{0,\dots,L\}$, where $c_l$ denotes the number of channels, $h_l$ is the height and $w_l$ is the width. The intermediate feature maps $\{E^l_t(\mathbf{x}_2)\}$, which are downsampled to the same spatial size as the output of the last convolutional block $E^L_t(\mathbf{x}_2)$ to reduce GPU memory consumption, are concatenated first and then mapped to a $d$-dimensional latent space through a $1\times1$ convolution followed by average pooling to obtain the ``\textbf{\textit{hypercolumn context}}'' $\mathbf{h}^h_2 \in \mathbb{R}^d$. $\mathbf{h}^h_2$ is transformed by another projector $H^h_t$ to obtain the hypercolumn embedding by $\mathbf{z}^h_2 = H^h_t(\mathbf{h}^h_2)$. Thus the contexts of the global feature $\mathbf{h}^g_2$ and hypercolumnn $\mathbf{h}^h_2$ are obtained for the teacher. Likewise, for the student, we produce the global feature context $\mathbf{h}^g_1 = E_s(\mathbf{x}_1)$, hypercolumnn context $\mathbf{h}^h_1$ and the corresponding embeddings $\mathbf{z}^g_1 = H_s(\mathbf{h}^g_1)$ and $\mathbf{z}^h_1 = H^h_s(\mathbf{h}^h_1)$ for the heavily corrupted view $\mathbf{x}_1$.

Next we measure the similarity relationships between the augmented views ($\mathbf{x}_1$ and $\mathbf{x}_2$) and the samples in the memory bank. Following~\cite{chen2020mocov2,zheng2021ressl}, the embeddings are used to maintain two queue-based memory banks separately: a memory bank $\mathcal{Q}$ based on $\mathbf{z}^g_2$ and a hypercolumn memory bank $\mathcal{Q}^h$ based on $\mathbf{z}^h_2$. To guide the learning of the global feature context $\mathbf{h}^g_1$ for the student, we use the similarity relationships between $\mathbf{h}^h_2$ and the embeddings $\hat{\mathbf{z}}^h_i$ in the hypercolumn memory bank $\mathcal{Q}^h$ as the target. The relationships are measured using the cosine similarity between $\mathbf{z}^h_2$ and $\hat{\mathbf{z}}^h_i$. We normalize the similarities with a softmax operation and produce a target probabilistic distribution $\mathbf{y}^h_2$ for the teacher:
\begin{equation}\label{eq:CGH-target}
	\mathbf{y}^h_2[i] = \frac{\exp{(\text{sim}{(\mathbf{z}^h_2, \hat{\mathbf{z}}^h_i) / \tau_h})}}{\sum_{k=1}^{M}{\exp{(\text{sim}{(\mathbf{z}^h_2, \hat{\mathbf{z}}^h_k)} / \tau_h)}}},
\end{equation}
where $\mathbf{y}^h_2[i]$ is the $i$-th element of the target similarity distribution generated by hypercolumn context $\mathbf{h}^h_2$, $\hat{\mathbf{z}}^h_i$ is the $i$-th embedding in the hypercolumn memory bank $\mathcal{Q}^h$, $\tau_h$ is the temperature parameter for the hypercolumn context, $M$ is the size of the memory bank and $\text{sim}(\mathbf{u}, \mathbf{v})=\frac{\mathbf{u}^\top\mathbf{v}}{{\lVert\mathbf{u}\rVert}_2{\lVert\mathbf{v}\rVert}_2}$ denotes the cosine similarity between the vectors $\mathbf{u}$ and $\mathbf{v}$. Similarly, the predicted distribution from the student is expressed as follows:
\begin{equation}\label{eq:CGH-predict}
	\mathbf{y}^g_1[i] = \frac{\exp{(\text{sim}{(\mathbf{z}^g_1, \hat{\mathbf{z}}_i / \tau_s)})}}{\sum_{k=1}^{M}{\exp{(\text{sim}{(\mathbf{z}^g_1, \hat{\mathbf{z}}_k)} / \tau_s)}}},
\end{equation}
where $\mathbf{y}^g_1[i]$ is the $i$-th element of the predicted similarity distribution generated by global feature context $\mathbf{h}^g_1$, $\hat{\mathbf{z}}_i$ is the $i$-th embedding in the memory bank $\mathcal{Q}$ and $\tau_s$ is the temperature for global feature context of the student. The global-hypercolumn alignment predicts the hypercolumn based similarity distribution $\mathbf{y}^h_2$ from the global feature based distribution $\mathbf{y}^g_1$ by minimizing the cross-entropy loss:
\begin{equation}\label{eq:CGH-gh}
	\mathcal{L}_\text{gh} = CE(\mathbf{y}^g_1, \mathbf{y}^h_2),
\end{equation}
where $CE(\mathbf{y}_1, \mathbf{y}_2) = -\sum_{k=1}^{M}{\mathbf{y}_2[k] \log{\mathbf{y}_1[k]}}$.

Similarly, for hypercolumn-global alignment, we guide the learning of the hypercolumn context $\mathbf{h}^h_1$ for the student using the global feature context $\mathbf{h}^g_2$ as target. The distributions generated by $\mathbf{h}^h_1$ and $\mathbf{h}^g_2$ are obtained as follows:
\begin{equation}\label{eq:CGH-hyper-def}
	\begin{array}{l}
		\mathbf{y}^h_1[i] = \frac{\exp{(\text{sim}{(\mathbf{z}^h_1, \hat{\mathbf{z}}^h_i / \tau_h)})}}{\sum_{k=1}^{M}{\exp{(\text{sim}{(\mathbf{z}^h_1, \hat{\mathbf{z}}^h_k)} / \tau_h)}}}, \\ 
		\mathbf{y}^g_2[i] = \frac{\exp{(\text{sim}{(\mathbf{z}^g_2, \hat{\mathbf{z}}_i) / \tau_t})}}{\sum_{k=1}^{M}{\exp{(\text{sim}{(\mathbf{z}^g_2, \hat{\mathbf{z}}_k)} / \tau_t)}}},
	\end{array}
\end{equation}
where $\tau_t$ is the temperature for the global feature context of the teacher. The objective for hypercolumn-global alignment is expressed as:
\begin{equation}\label{eq:CGH-hg}
	\mathcal{L}_\text{hg} = CE(\mathbf{y}^h_1, \mathbf{y}^g_2).
\end{equation}

Altogether, we enforce the cross-context learning between the global feature context and hypercolumn context with the following objective:
\begin{equation}\label{eq:CGH-global}
	\mathcal{L} = \mathcal{L}_\text{gh} + \mathcal{L}_\text{hg} = CE(\mathbf{y}^g_1, \mathbf{y}^h_2) + CE(\mathbf{y}^h_1, \mathbf{y}^g_2).
\end{equation}

\subsection{Momentum update}
The teacher is updated by the exponential moving average of the student:
\begin{equation}\label{eq:CGH-momentum}
	\begin{array}{l}
		E_t \leftarrow mE_t + (1-m)E_s, \\ 
		H_t \leftarrow mH_t + (1-m)H_s, \\ 
		H^h_t \leftarrow mH^h_t + (1-m)H^h_s,
	\end{array}
\end{equation}
where $m$ is the momentum coefficient, which is set to $0.999$ in all experiments following~\cite{chen2020mocov2,zheng2021ressl}.

\subsection{Architecture}
Following the common settings in self-supervised representation learning with Siamese structure~\cite{chen2020mocov2,grill2020bootstrap}, we use ResNet as the online encoder and its momentum-updated version as the momentum encoder. In our framework, the momentum encoder is used as the teacher and the online encoder is used as the student. As in~\cite{chen2020mocov2,zheng2021ressl}, a two-layer MLP is adopted as the projector $H_s$ for transforming the global feature from the global average pooling layer. Additionally, we adopt another two-layer MLP, which has the same architecture as $H_s$, as the projector $H^h_s$ for transforming the hypercolumn. Both projectors consist of two linear layers with a ReLU non-linear activation in between. Following ReSSL~\cite{zheng2021ressl}, the hidden and output dimension of both projectors are set to $4096$ and $512$, respectively. When transforming the concatenation of feature maps to generate the hypercolumn vector, we use a $1\times1$ convolutional layer followed by Batch Normalization (BN)~\cite{ioffe2015batch}, ReLU activation and global average pooling. In our experiments, we use the outputs of the four convolutional blocks of ResNet as intermediate feature maps.

\begin{table*}[htb]
	\caption{\textbf{Linear and KNN evaluation results on IN-1K with ResNet-50 backbone}. All methods are evaluated with the single-crop setting. Top-1 and Top-5 validation accuracy are reported. $^\dag$: our reproduction using the official codes. $\ast$: results cited from~\protect\cite{chen2020exploring}.}
	\centering
	\label{tab:CGH-ImageNet}
	\begin{tabular}{l c c c c c}
		\toprule
		Method & Backprop & Epochs & Batch Size & \thead{Linear \\ Acc.} & \thead{KNN \\ Acc.} \\
		\midrule
		Supervised & 1x & 100 & 256 & 76.5 & - \\
		\midrule
		\multicolumn{6}{l}{\textbf{Asymmetric loss.}} \\
		MoCo-v2~\cite{chen2020mocov2} & 1x & 200 & 256 & 67.5 & 55.9 \\
		PCL-v2~\cite{li2021prototypical} & 1x & 200 & 256 & 67.6 & 58.1 \\
		HCSC~\cite{guo2022hcsc} & 1x & 200 & 256 & 69.2 & 60.7 \\
		OBoW~\cite{Gidaris_2021_CVPR}$^\dag$ & 1x & 200 & 256 & 69.5 & 57.2 \\
		ReSSL~\cite{zheng2021ressl}$^\dag$ & 1x & 200 & 256 & 69.3 & 61.3 \\
		ReSSL-pred~\cite{zheng2021resslv2} & 1x & 200 & 1024 & 72.0 & - \\
		\rowcolor{WhiteSmoke!70!Lavender} CGH & 1x & 200 & 256 & \textbf{70.5} & \textbf{62.9} \\
		\rowcolor{WhiteSmoke!70!Lavender} CGH-pred & 1x & 200 & 256 & \textbf{72.3} & \textbf{65.8} \\
		\midrule
		\multicolumn{6}{l}{\textbf{Symmetric loss. $\mathbf{2\times}$ FLOPS}} \\
		SimCLR~\cite{pmlr-v119-chen20j}$\ast$ & 2x & 200 & 4096 & 68.3 & - \\
		SwAV~\cite{caron2020unsupervised}$\ast$ & 2x & 200 & 4096 & 69.1 & - \\
		SimSiam~\cite{chen2020exploring}$\ast$ & 2x & 200 & 256 & 70.0 & - \\
		BYOL~\cite{grill2020bootstrap}$\ast$ & 2x & 200 & 4096 & 70.6 & - \\
		NNCLR~\cite{dwibedi2021little} & 2x & 200 & 4096 & 70.7 & - \\
		\bottomrule
	\end{tabular}
\end{table*}

\begin{table*}[htb]
	\caption{Linear evaluation on IN-1K with multi-crop strategy~\cite{caron2020unsupervised,hu2021adco} and different pre-training epochs.}
	\centering
	\label{tab:CGH-multi}
	\begin{tabular}{l c c c c c c}
		\toprule
		Method & Backprop & Multi-Crop & Epochs & Batch Size & \thead{Linear \\ Acc.} \\
		\midrule
		CMSF~\cite{navaneet2022constrained} & 1x & \checkmark & 200 & 256 & 74.4 \\
		OBoW~\cite{Gidaris_2021_CVPR} & 1x & \checkmark & 200 & 256 & 73.8 \\
		ReSSL~\cite{zheng2021ressl} & 1x & \checkmark & 200 & 256 & 74.7 \\
		\rowcolor{WhiteSmoke!70!Lavender} CGH-pred & 1x & \checkmark & 200 & 256 & \textbf{75.7} \\
		\midrule
		SwAV~\cite{caron2020unsupervised} & 2x & \checkmark & 800 & 4096 & 75.3 \\
		HCSC~\cite{guo2022hcsc} & 1x & \checkmark & 800 & 256 & 74.2 \\
		CsMl~\cite{xu2022seed} & 2x & \checkmark & 300 & 1024 & 75.3 \\
		DINO~\cite{caron2021emerging} & 1x & \checkmark & 800 & 4096 & 75.3 \\
		NNCLR~\cite{dwibedi2021little} & 2x & $\times$ & 1000 & 4096 & 75.4 \\
		MAST~\cite{huang2023mast} & 2x & $\times$ & 1000 & 2048 & 75.8 \\
		\rowcolor{WhiteSmoke!70!Lavender} CGH-pred & 1x & \checkmark & 400 & 256 & \textbf{76.0} \\
		\bottomrule
	\end{tabular}
\end{table*}

\begin{table*}[htb]
	\caption{\textbf{IN-1K semi-supervised classification using ResNet-50 pre-trained on IN-1K}. \textbf{Multi} denotes the results with multi-crop. Top-1 and Top-5 validation accuracy are reported. $^\dag$: our reproduction using the official codes. $\ast$: results cited from~\protect\cite{huang2022learning}.}
	\centering
	\label{tab:CGH-semi}
	\begin{tabular}{l c c c c c c}
		\toprule
		\multirow{2}{*}{Method} & \multirow{2}{*}{Epochs} & \multirow{2}{*}{Batch Size} & \multicolumn{2}{c}{1\% Labels} & \multicolumn{2}{c}{10\% Labels} \\
		\cmidrule(lr){4-5} \cmidrule(lr){6-7}
		& & & Top-1 & Top-5 & Top-1 & Top-5 \\
		\midrule
		\multicolumn{7}{l}{\textbf{Asymmetric loss.}} \\
		MoCo-v2~\cite{chen2020mocov2}$\ast$ & 200 & 256 & 43.8 & 72.3 & 61.9 & 84.6 \\
		HCSC~\cite{guo2022hcsc} & 200 & 256 & 48.0 & 75.6 & 64.3 & 86.0 \\
		ReSSL~\cite{zheng2021ressl}$^\dag$ & 200 & 256 & 51.1 & 77.3 & 65.0 & 87.1 \\
		\rowcolor{WhiteSmoke!70!Lavender} CGH & 200 & 256 & \textbf{53.2} & \textbf{78.9} & \textbf{66.4} & \textbf{88.0} \\
		\midrule
		\multicolumn{7}{l}{\textbf{Symmetric loss. $\mathbf{2\times}$ FLOPS}} \\
		SimCLR~\cite{pmlr-v119-chen20j} & 1000 & 4096 & 48.3 & 75.5 & 65.6 & 87.8 \\
		SwAV~\cite{caron2020unsupervised} & 800 & 4096 & 53.9 & 78.5 & 70.2 & 89.9 \\
		BYOL~\cite{grill2020bootstrap} & 1000 & 4096 & 53.2 & 78.4 & 68.8 & 89.0 \\
		\midrule
		\multicolumn{7}{l}{\textbf{Multi-crop}} \\
		ReSSL (Multi)~\cite{zheng2021ressl} & 200 & 256 & 57.9 & - & \textbf{70.4} & - \\
		\rowcolor{WhiteSmoke!70!Lavender} CGH (Multi) & 200 & 256 & \textbf{58.4} & \textbf{82.4} & 70.3 & \textbf{90.3} \\
		\bottomrule
	\end{tabular}
\end{table*}

\begin{table}[htb]
	\caption{\textbf{Transfer learning on PASCAL VOC object detection}. All models are pre-trained for 200 epochs on IN-1K using ResNet-50 as the encoder. ResNet-50-C4 is used as the fine-tuning backbone. The bounding-box detection score ($\text{AP}^\text{bb}$) is reported. $^\dag$: our reproduction using the official codes. $\ast$: results cited from~\protect\cite{chen2020exploring}.}
	\centering
	\label{tab:CGH-detection}
	\begin{tabular}{l c c c}
		\toprule
		Method & $\text{AP}^\text{bb}$ & $\text{AP}^\text{bb}_{50}$ & $\text{AP}^\text{bb}_{75}$ \\
		\midrule
		\multicolumn{4}{l}{\textbf{Asymmetric loss.}} \\
		MoCo-v2~\cite{chen2020mocov2} & 57.0 & 82.4 & 63.6 \\
		ReSSL~\cite{zheng2021ressl}$^\dag$ & 56.1 & 82.2 & 62.5 \\
		\rowcolor{WhiteSmoke!70!Lavender} CGH & 56.8 & \textbf{82.6} & 63.4 \\
		\rowcolor{WhiteSmoke!70!Lavender} CGH (Multi) & \textbf{57.1} & \textbf{82.6} & \textbf{63.8} \\
		\midrule
		\multicolumn{4}{l}{\textbf{Symmetric loss. $\mathbf{2\times}$ FLOPS}} \\
		SimCLR~\cite{pmlr-v119-chen20j}$\ast$ & 55.5 & 81.8 & 61.4 \\
		SwAV~\cite{caron2020unsupervised}$\ast$ & 55.4 & 81.5 & 61.4 \\
		SimSiam~\cite{chen2020exploring}$\ast$ & 56.4 & 82.0 & 62.8 \\
		BYOL~\cite{grill2020bootstrap}$\ast$ & 55.3 & 81.4 & 61.1 \\
		\bottomrule
	\end{tabular}
\end{table}

\section{Experiments}
In this section, we perform performance evaluation on widely used self-supervised learning benchmarks, including classification dataset ImageNet-1k~\cite{deng2009imagenet} (also known as IN-1K) and detection datasets (i.e, PASCAL VOC~\cite{everingham2010pascal} and COCO~\cite{lin2014microsoft}). The visualization results are provided in the supplementary material.

\subsection{Experimental setups}
\subsubsection{Implementation details}

We adopt the same encoder backbone for all methods. The experiments on non-ImageNet datasets adopt ResNet-18 while the experiments on IN-1K use ResNet-50. For contrastive augmentation, we use the same strategy as in contrastive learning~\cite{chen2020mocov2}. For weak augmentation, we use random resized crop and random horizontal flip, which is also the practice in ReSSL.

The teacher temperature and student temperature for global feature are set to $\tau_t=0.04$ and $\tau_s=0.1$ respectively, following ReSSL. The hypercolumn temperature is set to $\tau_h=0.08$. The outputs of the third and fourth convolutional block are used for generating the hypercolumn. For a fair comparison, the other hyper-parameters are kept the same as ReSSL in all experiments. In the revised version of ReSSL~\cite{zheng2021resslv2}, an additional predictor is used to further improve performance (denoted as \textbf{ReSSL-pred}). We also report our results with a predictor (denoted as \textbf{CGH-pred}) using the same pre-training details as discussed above. Note that \textbf{2x backprop} methods update the encoder parameters twice at each training step using the two augmented views, which means more samples are used within the same epochs and much higher training cost than \textbf{1x backprop} methods~\cite{zheng2021ressl,huang2022learning} like ours.

\subsubsection{Training details}
By default, the pre-training is performed on the training set of IN-1K with 2 NVIDIA A100 GPUs. In ablation studies, we perform the pre-training on Tiny-ImageNet~\cite{Le2015TinyIV} and STL-10~\cite{pmlr-v15-coates11a} for $400$ epochs. The training recipes are detailed as follows.

\textbf{Tiny-ImageNet/STL-10}. Following ReSSL, we pre-train for 400 epochs, using the SGD optimizer with 0.06 learning rate, 5e-4 weight decay, and 0.9 momentum. The batch size is set to 256. Following the linear evaluation protocol in~\cite{kim2020mixco}, we train the classifier for 100 epochs with a batch size of 256, 3.0 learning rate, no weight decay, 0.9 momentum and cosine learning rate decay.

\textbf{IN-1K}. Following ReSSL, we pre-train the model for 200 epochs, using the SGD optimizer with 0.05 learning rate, 1e-4 weight decay, and 0.9 momentum. The batch size is set to 256. As in ReSSL, for linear evaluation, we use 0.3 learning rate, no weight decay, 0.9 momentum and cosine learning rate decay.

\subsection{Linear classification and KNN evaluation}
In this section, following the linear evaluation protocol~\cite{chen2020mocov2,pmlr-v119-chen20j}, we evaluate the learned representations by learning a linear classifier on top of the frozen pre-trained encoder for classification task. The encoder is pre-trained on the training set of the dataset first and then the linear classifier is trained on the training set with labels. The classification accuracy on the validation set is reported. For KNN evaluation, we follow the protocol in~\cite{wu2018unsupervised,guo2022hcsc} by evaluating the learned encoder with K-nearest neighbor (KNN) classifier using several nearest neighbor settings $\{10, 20, 100, 200\}$ and reporting the highest accuracy.

The linear and KNN classification results on IN-1K using 200 pre-training epochs are provided in~\cref{tab:CGH-ImageNet}. The proposed method outperforms MoCo-v2/ReSSL by $3.0\%$/$1.2\%$ on linear classification and $7.0\%$/$1.6\%$ on KNN classification, respectively. The consistent improvement compared with the baselines shows the effectiveness of the proposed cross-context learning strategy.

To further demonstrate our performance, we provide the classification results on IN-1K with multi-crop strategy~\cite{hu2021adco} and longer pre-training epochs in~\cref{tab:CGH-multi}. As we can see, the proposed CGH outperforms previous state-of-the-art methods. Note that our method also outperforms strong baselines that learn from multi-level signals (intermediate features), such as OBoW~\cite{Gidaris_2021_CVPR} and CsMl~\cite{xu2022seed}. \textbf{The difference between our CGH and these multi-level methods are discussed in the supplementary material}.

\subsection{Semi-supervised classification}
We report the semi-supervised learning results by fine-tuning the self-supervised pre-trained ResNet-50 using $1\%$ and $10\%$ labelled data in IN-1K. We follow the semi-supervised protocol of~\cite{zheng2021ressl,pmlr-v119-chen20j} and report the results in~\cref{tab:CGH-semi}. We use SGD optimizer with batch size of $256$, weight decay of $0$, and momentum of $0.9$ for fine-tuning. For $1\%$ setting, we train for 50 epochs using initial learning rate of $0.5$ and $0.0001$ for the classification head and feature extractor backbone, respectively, which are decayed by a factor of $0.1$ after 30 and 40 epochs. In the $10\%$ setting, we fine-tune for 50 epochs and set the initial learning rate to $0.2$ and $0.0002$ for the classification head and feature extractor backbone, respectively, which are decayed by a factor of $0.1$ at the 30-th and 40-th epoch. Our method outperforms the other methods significantly with 200 pre-training epochs. Moreover, we also report the results with multi-crop in the last section of~\cref{tab:CGH-semi}. In this case, our method achieves better performance than ReSSL on $1\%$ split and comparable results on $10\%$ split. Furthermore, our method outperforms 2x backprop methods with more pre-training epochs.

\subsection{Transfer learning}
We evaluate the transfer learning performance of the learned representations on the object detection and instance segmentation task. We fine-tune the model pre-trained on IN-1K on two widely used benchmarks PASCAL VOC~\cite{everingham2010pascal} and COCO~\cite{lin2014microsoft}. The same protocol and setups as MoCo-v2 are adopted. For PASCAL VOC object detection, we adopt Faster R-CNN~\cite{ren2015faster} as the detector backbone, which is fine-tuned on training and validation splits of VOC 2007 and VOC 2012 and then tested on test set of VOC 2007; for COCO detection and segmentation, we use the Mask R-CNN~\cite{he2017mask} backbone, which is trained on the training set and then evaluated on the validation set. The results on PASCAL VOC are reported in~\cref{tab:CGH-detection} while the performance on COCO can be found in the supplementary material. As we can see, the proposed method achieves competitive performance compared with the state-of-the-art methods, which demonstrates the generality of the learned representations.

\begin{table}[htb]
	\caption{Comparison of different context variants.}
	\centering
	\label{tab:CGH-context}
	\begin{tabular}{l c c}
		\toprule
		Method & Tiny-ImageNet & STL-10 \\
		\midrule
		CGH (global-context) & 48.9 & 90.7 \\
		CGH (same-context) & 51.6 & 91.0 \\
		CGH (cross-context) & \textbf{53.8} & \textbf{92.0} \\
		\bottomrule
	\end{tabular}
\end{table}

\begin{table}[htb]
	\caption{Effect of hypercolumn temperature on Tiny-ImageNet.}
	\centering
	\label{tab:CGH-temperature}
	\begin{tabular}{l c c c c c}
		\toprule
		$\tau_h$ & $0.02$ & $0.04$ & $0.06$ & $0.08$ & $0.1$ \\
		\midrule
		Acc. & 52.1 & 52.8 & 53.1 & \textbf{53.8} & 51.8 \\
		\bottomrule
	\end{tabular}
\end{table}

\begin{table}[htb]
	\caption{Effect of combinations of intermediate layers.}
	\centering
	\label{tab:CGH-layer}
	\begin{tabular}{c c c c c c}
		\toprule
		\multicolumn{4}{c}{Layers} & \multirow{2}{*}{Tiny-ImageNet} & \multirow{2}{*}{STL-10} \\
		\cmidrule{1-4}
		L1 & L2 & L3 & L4 \\
		\midrule
		- & - & - & \checkmark & 48.9 & 90.7 \\
		\checkmark & - & - & \checkmark & 53.6 & 91.8 \\
		- & \checkmark & - & \checkmark & 54.1 & 92.2 \\
		- & - & \checkmark & \checkmark & 53.8 & 92.0 \\
		\checkmark & \checkmark & \checkmark & \checkmark & \textbf{54.4} & \textbf{92.3} \\
		\bottomrule
	\end{tabular}
\end{table}

\subsection{Ablation studies}
\subsubsection{Comparison of different context variants}
In contrast to global-context learning, we leverage the similarity relationships of one context (\eg, hypercolumn) as a supervisory signal for the other context (\eg, global feature). Alternatively, in addition to the consistency of the global features used in global-context learning framework, we could incorporate the context of hypercolumn by enforcing the consistency of the hypercolumns between the teacher and the student, which is termed as same-context. We compare global-context, same-context and cross-context in~\cref{tab:CGH-context}. Note that global-context is identical to ReSSL here. We find that by incorporating the context from intermediate layers, both same-context and cross-context outperform the global-context baseline. Moreover, cross-context achieves the best results. \textbf{This suggests that cross-context provides a superior supervisory signal compared with global-context learning because of the use of the other contexts for supervision}.

\subsubsection{Hypercolumn temperature}
We use a temperature to control the smoothness of the generated similarity distribution. Therefore, the temperature is an important hyper-parameter in our framework. In order to evaluate the effect of the temperature, we evaluate the values of $\tau_h$ from set $\{0.02, 0.04, 0.06, 0.08, 0.1\}$. As shown in~\cref{tab:CGH-temperature}, we observe an inverted U-shaped trend on the performance when we increase $\tau_h$.

Note that when $\tau_h \rightarrow 0$, the distribution from the hypercolumn becomes extremely sharp. If it is used for target generation, the target turns into one-hot distribution where the goal is to match the query with the most similar sample from the memory bank instead of capturing the instance relations. In other words, the framework degrades to NNCLR~\cite{dwibedi2021little}, except NNCLR doesn't use hypercolumn or memory bank for generating candidate samples. By contrast, when $\tau_h \rightarrow 0.1$, the distribution becomes flat and fails to focus on similar samples. Therefore, the performance tends to be better when $\tau_h$ is within $\left[ {0.06, 0.08} \right]$.

\subsubsection{Intermediate layers for hypercolumn}
It is interesting to explore the effectiveness and sensitivity of different combinations of the intermediate layers. We use one of the layers from layer1 to layer3, along with the layer4 to generate hypercolumn and analyze the effect. The results are provided in~\cref{tab:CGH-layer}. We have the following observations: 1) The proposed CGH achieves the best result by using all four layers for hypercolumn generation. 2) Hypercolumn based on layer2 and layer4 achieves the second best result, which suggests that layer2 provides a better balance between low-level and high-level semantics compared with layer1 and layer3. 3) Regardless of different combinations of the intermediate layers, all variants outperform the baseline ReSSL (first row), which shows our method is robust to the choice of the intermediate layers. Note that when we only use the fourth convolutional block for hypercolumn generation, our model will be identical to ReSSL -- this is the first row of~\cref{tab:CGH-layer}.

\begin{figure}[htb]
	\centering
	\includegraphics[width=0.9\linewidth]{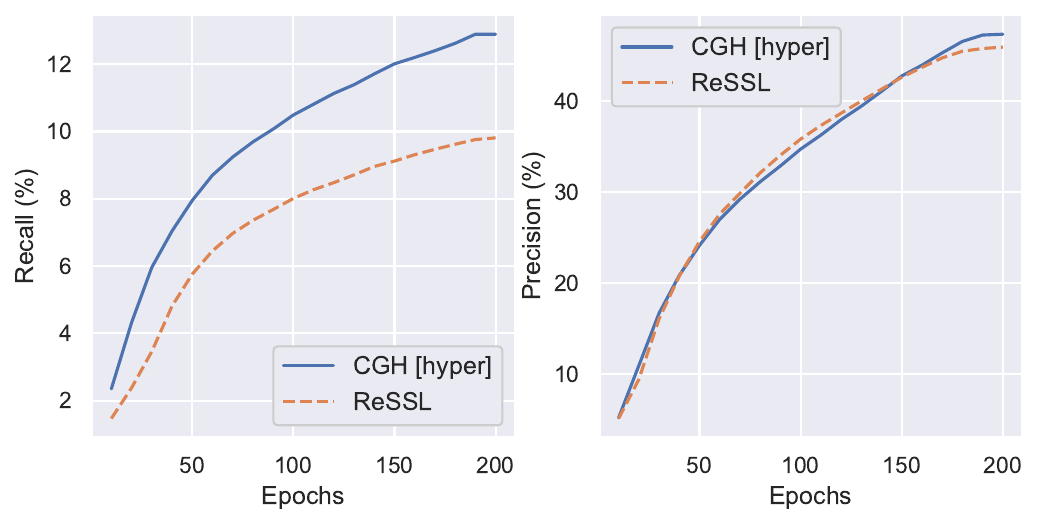}
	\caption{Performance of positive sample selection on IN-1K.}
	\label{fig:CGH-precision/recall}
\end{figure}

\subsection{Effectiveness of hypercolumn context}\label{sec:CGH-effect}
In this section, we demonstrate the benefits of using hypercolumn representations for learning global representations from the perspective of (soft-) selection of positive samples, following the protocol in~\cite{guo2022hcsc}. To do so, we note that learning the student under the guidance of the teacher can be viewed as training with soft pseudo-labels provided by the teacher. By thresholding the similarity distribution of the teacher, the predicted positives and negatives in the memory bank are obtained. Since the labels of the dataset are publicly available, the ground-truth for the positives and negatives in the memory bank can also be obtained (the sample is positive if it belongs to the same class as the input). Therefore, we can calculate the recall and precision which indicate the true positive and false positive selected by the teacher's similarity distribution. In~\cref{fig:CGH-precision/recall}, we provide the plots of recall and precision on IN-1K during pre-training based on the hupercolumn distribution from the teacher, and the corresponding plots of ReSSL based on global-context distribution. It's shown that our method has considerably better recall than ReSSL for the duration of the training while maintaining similar precision levels (right hand side plot). In summary, the results show that the proposed scheme can find more correct positive samples corresponding to the same class as the input (true positives), and maintain a low false positive rate at the same time.

\section{Conclusion}
In order to solve the class collision problem in contrastive learning, inspired by cross-modality learning~\cite{alwassel2020self,peng2022balanced}, we present a novel framework based on knowledge distillation, cross-context learning between global and hypercolumn features (CGH) that learns representations by capturing cross-context information from the context of global features and hypercolumns. The cross-context learning strategy allows the model to identify more similar samples (true positives) in the memory bank and keep low false positives. The extensive experiments on classification and downstream tasks demonstrate the effectiveness and generality of our method.

\section*{Acknowledgement}
This work was supported by the EU H2020 AI4Media No.951911 project. We thank James Oldfield for his helpful comments.

{\small
	\bibliographystyle{ieee_fullname}
	\bibliography{references}
}

\clearpage

\section*{Supplementary Material}

\appendix

\section{Discussion with works related to intermediate features}
\begin{figure}[htb]
	\centering
	\includegraphics[width=0.9\linewidth]{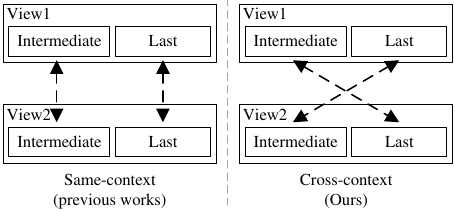}
	\caption{Comparison of our method with previous works.}
	\label{fig:CGH-compare}
\end{figure}

While some existing works have already explored multi-level self-supervisions (intermediate features) in self-supervised learning (SSL)~\cite{Gidaris_2021_CVPR,xu2022seed,yoshihashi2022ladder,jang2023self}, our method is fundamentally different from these works in the following aspects:
\begin{itemize}
	\item \textbf{Our goal is to present a novel framework to alleviate the ``class collision problem'' in contrastive learning~\cite{chen2022perfectly,li2021prototypical}}, which is different from the goal of previous works where they aim to apply self-supervised objective over intermediate levels.
	\item \textbf{Previous works commonly adopt a straightforward way to extend the self-supervised objective over the last global features to multi-level learning on the intermediate features} by inducing both intermediate features and global features to the self-supervised objective simultaneously~\cite{Gidaris_2021_CVPR,xu2022seed,yoshihashi2022ladder}. In practice, most of them use the intermediate features from the teacher to supervise the corresponding features from the student (\textcolor{red}{same-context in~\cref{fig:CGH-compare}}), which is the application of knowledge distillation in SSL. \textbf{In contrast, we propose a cross-layer learning strategy where intermediate features and global features are used as each other's supervisory signal (\textcolor{red}{cross-contest in~\cref{fig:CGH-compare})}}. The superiority of cross-context over same-context is shown in ablation (\textcolor{red}{Tab. 5 in the main paper}). \textbf{Moreover, we outperform OBoW~\cite{Gidaris_2021_CVPR}, which also adopts the same-context strategy (\textcolor{red}{Tab. 1, 2 in the main paper,~\cref{tab:CGH-detection-coco} in the supplementary material})}.
	\item Another work~\cite{jang2023self} encourages the intermediate representations to learn from the last layer via the contrastive loss, which is still different from our cross-context (cross-layer) learning. Besides, our objective measures the instance relations with cross-entropy loss to alleviate the ``class collision problem'' while work~\cite{jang2023self} fails to do so as it is still based on contrastive objective.
	\item \textbf{Therefore, compared with current works, we have a different goal and to achieve that goal we adopt a different way of leveraging intermediate features for producing better supervisory signal}.
\end{itemize}

\section{Additional experiment results}

\begin{table}[htb]
	\caption{\textbf{Results of IN-1K linear classification using hypercolumn}. \textbf{hyper} is the result using hypercolumn as the input to the linear classifier.}
	\centering
	\label{tab:cgh-hyper}
	\begin{tabular}{l c}
		\toprule
		Method & IN-1K Acc. \\
		\midrule
		MoCo-v2 & 67.5 \\
		ReSSL & 69.3 \\
		CGH & 70.5 \\
		CGH (hyper) & \textbf{70.8} \\
		\bottomrule
	\end{tabular}
\end{table}

\begin{figure}[htb]
	\centering
	\includegraphics[width=0.8\linewidth]{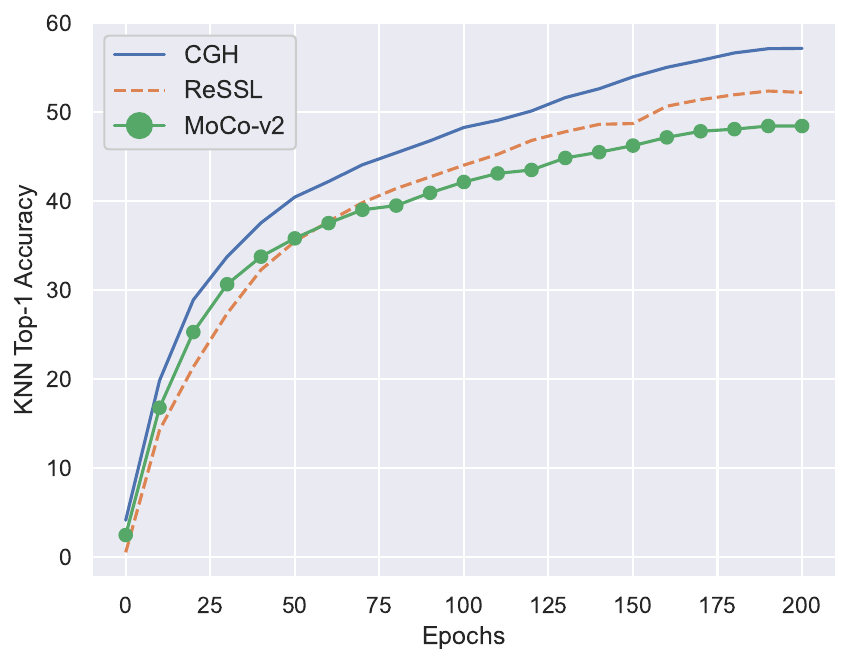}
	\caption{KNN validation accuracy with respect to epochs during pre-training.}
	\label{fig:CGH-knn}
\end{figure}

\begin{table*}[htb]
	\caption{\textbf{Transfer learning on COCO object detection and instance segmentation using ResNet-50 pre-trained on IN-1K}. We report the bounding-box AP ($\text{AP}^\text{bb}$) for object detection and mask AP ($\text{AP}^\text{mk}$) for instance segmentation. $^\dag$: our reproduction using the official codes. $\ast$: results cited from~\protect\cite{chen2020exploring}.}
	\centering
	\label{tab:CGH-detection-coco}
	\begin{tabular}{l c c c c c c c}
		\toprule
		\multirow{2}{*}{Method} & \multirow{2}{*}{Epochs} & \multicolumn{3}{c}{COCO Det.} & \multicolumn{3}{c}{COCO Instance Seg.} \\
		\cmidrule(lr){3-5} \cmidrule(lr){6-8}
		& & $\text{AP}^\text{bb}$ & $\text{AP}^\text{bb}_{50}$ & $\text{AP}^\text{bb}_{75}$ & $\text{AP}^\text{mk}$ & $\text{AP}^\text{mk}_{50}$ & $\text{AP}^\text{mk}_{75}$ \\
		\midrule
		\multicolumn{7}{l}{\textbf{Asymmetric loss.}} \\
		MoCo-v2~\cite{chen2020mocov2} & 200 & 38.8 & 58.0 & 42.0 & 34.0 & 55.2 & 36.3 \\
		OBoW~\cite{Gidaris_2021_CVPR}$^\dag$ & 200 & 38.6 & 58.0 & 41.8 & 33.8 & 54.8 & 36.2 \\
		ReSSL~\cite{zheng2021ressl}$^\dag$ & 200 & 38.3 & 57.7 & 41.3 & 33.4 & 54.7 & 35.3 \\
		\rowcolor{WhiteSmoke!70!Lavender} CGH & 200 & \textbf{39.0} & \textbf{58.8} & \textbf{42.2} & \textbf{34.2} & \textbf{55.3} & \textbf{36.5} \\
		\midrule
		\multicolumn{7}{l}{\textbf{Symmetric loss. $\mathbf{2\times}$ FLOPS}} \\
		SimCLR~\cite{pmlr-v119-chen20j}$\ast$ & 200 & 37.9 & 57.7 & 40.9 & 33.3 & 54.6 & 35.3 \\
		SwAV~\cite{caron2020unsupervised}$\ast$ & 200 & 37.6 & 57.6 & 40.3 & 33.1 & 54.2 & 35.1 \\
		SimSiam~\cite{chen2020exploring}$\ast$ & 200 & 37.9 & 57.5 & 40.9 & 33.2 & 54.2 & 35.2 \\
		BYOL~\cite{grill2020bootstrap}$\ast$ & 200 & 37.9 & 57.8 & 40.9 & 33.2 & 54.3 & 35.0 \\
		LEWEL~\cite{huang2022learning} & 200 & 38.5 & 58.9 & 41.2 & 33.7 & 55.5 & 35.5 \\
		\midrule
		\multicolumn{8}{l}{\textbf{Multi-crop}} \\
		\rowcolor{WhiteSmoke!70!Lavender} CGH (Multi) & 200 & \textbf{39.3} & \textbf{59.3} & \textbf{42.7} & \textbf{34.4} & \textbf{55.9} & \textbf{36.6} \\
		\bottomrule
	\end{tabular}
\end{table*}

\subsection{IN-1K classification using hypercolumn}
We investigate the effectiveness of hypercolumn by using it for linear classification in~\cref{tab:cgh-hyper}. CGH with hypercolumn outperforms its counterpart that directly uses representation vectors after global average pooling for classification, which indicates the hypercolumn provides better supervisory signal.

\begin{figure*}[htb]
	\centering
	\includegraphics[width=0.9\linewidth]{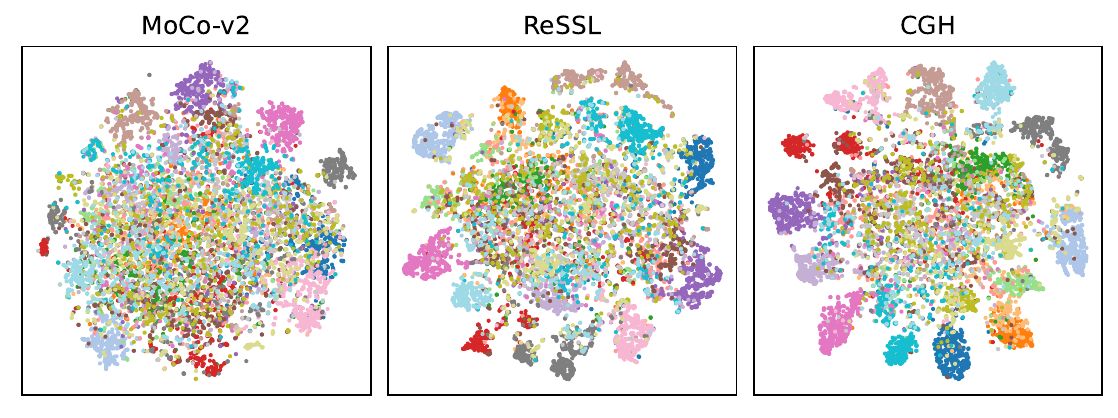}
	\caption{The t-SNE visualization on the training set of Tiny-ImageNet for the first 20 classes. The classes are represented by different colours.}
	\label{fig:CGH-tsne}
\end{figure*}

\begin{table}[htb]
	\caption{Comparison of pre-training running time relative to supervised training.}
	\centering
	\label{tab:CGH-time}
	\begin{tabular}{l c c c}
		\toprule
		Method & \thead{Time/\\ Epoch} & \thead{Linear \\ Acc.} & \thead{VOC 07+12 \\ Det.} \\
		\midrule
		Supervised & 1.00 & 76.5 & 81.3 \\
		MoCo-v2~\cite{chen2020mocov2} & 1.62 & 67.5 & 82.4 \\
		ReSSL~\cite{zheng2021ressl} & 1.62 & 69.3 & 82.2 \\
		BYOL~\cite{grill2020bootstrap} & 2.90 & 70.6 & 81.4 \\
		\rowcolor{WhiteSmoke!70!Lavender} CGH & 2.01 & 70.5 & 82.6 \\
		\bottomrule
	\end{tabular}
\end{table}

\subsection{Visualization of training progress}
Following~\cite{wu2018unsupervised}, we present the KNN classification accuracy with respect to epoch number in~\cref{fig:CGH-knn}, which is a useful metric to monitor the training progress. The KNN classifier is evaluated on the validation set of IN-1K. The KNN accuracy plot shows that the proposed method achieves a steady and consistent improvement. Note that in~\cref{fig:CGH-knn} we perform the KNN classification using the embedding from the MLP head as in~\cite{wu2018unsupervised}. However, in KNN evaluation, we build the KNN classifier on top of the global average pooling layer of ResNet by following~\cite{guo2022hcsc}.

\subsection{COCO object detection and instance segmentation}\label{sec:CGH-coco}
For COCO object detection and instance segmentation, we fine-tune the Mask R-CNN~\cite{he2017mask} with ResNet-50-C4 backbone using the model pre-trained on IN-1K. Following~\cite{chen2020mocov2,zheng2021ressl}, we adopt the 1x schedule used in the detetron2~\cite{wu2019detectron2}, which fine-tunes the model for $90,000$ iterations. The results on COCO are reported in~\cref{tab:CGH-detection-coco}. CGH outperforms ReSSL on all tasks, which demonstrates the effectiveness of the learned representations. Moreover, our method achieves better performance than 2x backprop methods like SimCLR, SwAV, SimSiam and BYOL and competitive results with SOTA methods like MoCo-v2~\cite{chen2020mocov2} and LEWEL~\cite{huang2022learning}.

\section{Visualization of feature representations}\label{sec:CGH-vis}
We use t-SNE~\cite{hinton2002stochastic} to visualize the learned representation on the training set of Tiny-ImageNet. The first 20 classes of Tiny-ImageNet are selected for the visualization. We report the comparison results of three methods, i.e., MoCo-v2, ReSSL and the proposed CGH in~\cref{fig:CGH-tsne}. As shown in~\cref{fig:CGH-tsne}, the proposed CGH has better class separation than MoCo-v2 and ReSSL. The t-SNE visualization results demonstrate that the proposed CGH can produce more discriminative representations, which benefit the performance on various downstream tasks.

\section{Training cost analysis}\label{sec:CGH-cost}
In this section, we compare the training cost of our method with the baselines in~\cref{tab:CGH-time}. For all methods, we perform the pre-training on IN-1K for 200 epochs with ResNet-50 backbone using 2 NVIDIA A100 GPUs. We measure the time consumption relative to supervised IN-1K training (``Supervised") based on the running time of one training epoch (``Time/Epoch"). Note that BYOL uses a batch size of 4096 to achieve the reported performance while we report the training cost using a batch size of 256 due to limited GPU memory. The results show that CGH outperforms ReSSL by $1.2\%$ and $0.4\%$ on IN-1K linear classification and PASCAL VOC object detection with reasonable cost increase ($2.01$ vs. $1.62$). Moreover, compared with 2x backprop methods like BYOL, the proposed method achieves $1.2\%$ improvement on detection and similar performance on classification ($70.5$ vs. $70.6$) with much less training cost ($2.01$ vs. $2.90$).

\section{Negative societal impact}
Generally self-supervised learning needs to pre-train with multiple GPUs for a long time to achieve competitive results with supervised learning. Our method also has such limitation. However, our method has better performance than SOTA self-supervised learning methods with similar (or shorter) training time, \eg, our CGH (1x backprop) achieves compatible performance with BYOL (2x backprop method with longer training time) on classification and object detection (\cref{tab:CGH-time}).

{\small
	\bibliographystyle{ieee_fullname}
	\bibliography{references}

\begin{thebibliography}{10}\itemsep=-1pt

\bibitem{alwassel2020self}
Humam Alwassel, Dhruv Mahajan, Bruno Korbar, Lorenzo Torresani, Bernard Ghanem,
  and Du Tran.
\newblock Self-supervised learning by cross-modal audio-video clustering.
\newblock In {\em Advances in Neural Information Processing Systems},
  volume~33, 2020.

\bibitem{caron2018deep}
Mathilde Caron, Piotr Bojanowski, Armand Joulin, and Matthijs Douze.
\newblock Deep clustering for unsupervised learning of visual features.
\newblock In {\em European Conference on Computer Vision}, 2018.

\bibitem{caron2020unsupervised}
Mathilde Caron, Ishan Misra, Julien Mairal, Priya Goyal, Piotr Bojanowski, and
  Armand Joulin.
\newblock Unsupervised learning of visual features by contrasting cluster
  assignments.
\newblock In {\em Proceedings of Advances in Neural Information Processing
  Systems (NeurIPS)}, 2020.

\bibitem{caron2021emerging}
Mathilde Caron, Hugo Touvron, Ishan Misra, Hervé Jegou, Julien Mairal, Piotr
  Bojanowski, and Armand Joulin.
\newblock Emerging properties in self-supervised vision transformers.
\newblock In {\em 2021 IEEE/CVF International Conference on Computer Vision
  (ICCV)}, pages 9630--9640, 2021.

\bibitem{chen2022perfectly}
Mayee Chen, Daniel~Y Fu, Avanika Narayan, Michael Zhang, Zhao Song, Kayvon
  Fatahalian, and Christopher Re.
\newblock Perfectly balanced: Improving transfer and robustness of supervised
  contrastive learning.
\newblock In Kamalika Chaudhuri, Stefanie Jegelka, Le Song, Csaba Szepesvari,
  Gang Niu, and Sivan Sabato, editors, {\em Proceedings of the 39th
  International Conference on Machine Learning}, volume 162 of {\em Proceedings
  of Machine Learning Research}, pages 3090--3122. PMLR, 17--23 Jul 2022.

\bibitem{pmlr-v119-chen20j}
Ting Chen, Simon Kornblith, Mohammad Norouzi, and Geoffrey Hinton.
\newblock A simple framework for contrastive learning of visual
  representations.
\newblock In Hal~Daumé III and Aarti Singh, editors, {\em Proceedings of the
  37th International Conference on Machine Learning}, volume 119 of {\em
  Proceedings of Machine Learning Research}, pages 1597--1607. PMLR, 13--18 Jul
  2020.

\bibitem{chen2020big}
Ting Chen, Simon Kornblith, Kevin Swersky, Mohammad Norouzi, and Geoffrey~E
  Hinton.
\newblock Big self-supervised models are strong semi-supervised learners.
\newblock In H. Larochelle, M. Ranzato, R. Hadsell, M.F. Balcan, and H. Lin,
  editors, {\em Advances in Neural Information Processing Systems}, volume~33,
  pages 22243--22255. Curran Associates, Inc., 2020.

\bibitem{chen2020mocov2}
Xinlei Chen, Haoqi Fan, Ross Girshick, and Kaiming He.
\newblock Improved baselines with momentum contrastive learning.
\newblock {\em arXiv preprint arXiv:2003.04297}, 2020.

\bibitem{chen2020exploring}
Xinlei Chen and Kaiming He.
\newblock Exploring simple siamese representation learning.
\newblock In {\em 2021 IEEE/CVF Conference on Computer Vision and Pattern
  Recognition (CVPR)}, pages 15745--15753, 2021.

\bibitem{pmlr-v15-coates11a}
Adam Coates, Andrew Ng, and Honglak Lee.
\newblock An analysis of single-layer networks in unsupervised feature
  learning.
\newblock In Geoffrey Gordon, David Dunson, and Miroslav Dudík, editors, {\em
  Proceedings of the Fourteenth International Conference on Artificial
  Intelligence and Statistics}, volume~15 of {\em Proceedings of Machine
  Learning Research}, pages 215--223, Fort Lauderdale, FL, USA, 11--13 Apr
  2011. PMLR.

\bibitem{deng2009imagenet}
J. {Deng}, W. {Dong}, R. {Socher}, L. {Li}, {Kai Li}, and {Li Fei-Fei}.
\newblock Imagenet: A large-scale hierarchical image database.
\newblock In {\em 2009 IEEE Conference on Computer Vision and Pattern
  Recognition}, pages 248--255, 2009.

\bibitem{dwibedi2021little}
Debidatta Dwibedi, Yusuf Aytar, Jonathan Tompson, Pierre Sermanet, and Andrew
  Zisserman.
\newblock With a little help from my friends: Nearest-neighbor contrastive
  learning of visual representations.
\newblock In {\em Proceedings of the IEEE/CVF International Conference on
  Computer Vision}, pages 9588--9597, 2021.

\bibitem{everingham2010pascal}
Mark Everingham, Luc Van~Gool, Christopher~KI Williams, John Winn, and Andrew
  Zisserman.
\newblock The pascal visual object classes (voc) challenge.
\newblock {\em International Journal of Computer Vision}, 88(2):303--338, 2010.

\bibitem{feng2022adaptive}
Chen Feng and Ioannis Patras.
\newblock Adaptive soft contrastive learning.
\newblock In {\em 2022 26th International Conference on Pattern Recognition
  (ICPR)}, pages 2721--2727. IEEE, 2022.

\bibitem{Gidaris_2021_CVPR}
Spyros Gidaris, Andrei Bursuc, Gilles Puy, Nikos Komodakis, Matthieu Cord, and
  Patrick Pérez.
\newblock Obow: Online bag-of-visual-words generation for self-supervised
  learning.
\newblock In {\em 2021 IEEE/CVF Conference on Computer Vision and Pattern
  Recognition (CVPR)}, pages 6826--6836, 2021.

\bibitem{grill2020bootstrap}
Jean-Bastien Grill, Florian Strub, Florent Altch\'{e}, Corentin Tallec, Pierre
  Richemond, Elena Buchatskaya, Carl Doersch, Bernardo Avila~Pires, Zhaohan
  Guo, Mohammad Gheshlaghi~Azar, Bilal Piot, koray kavukcuoglu, Remi Munos, and
  Michal Valko.
\newblock Bootstrap your own latent - a new approach to self-supervised
  learning.
\newblock In H. Larochelle, M. Ranzato, R. Hadsell, M.F. Balcan, and H. Lin,
  editors, {\em Advances in Neural Information Processing Systems}, volume~33,
  pages 21271--21284. Curran Associates, Inc., 2020.

\bibitem{guo2022hcsc}
Yuanfan Guo, Minghao Xu, Jiawen Li, Bingbing Ni, Xuanyu Zhu, Zhenbang Sun, and
  Yi Xu.
\newblock Hcsc: Hierarchical contrastive selective coding.
\newblock In {\em 2022 IEEE/CVF Conference on Computer Vision and Pattern
  Recognition (CVPR)}, pages 9696--9705, 2022.

\bibitem{hariharan2015hypercolumns}
Bharath Hariharan, Pablo Arbeláez, Ross Girshick, and Jitendra Malik.
\newblock Hypercolumns for object segmentation and fine-grained localization.
\newblock In {\em 2015 IEEE Conference on Computer Vision and Pattern
  Recognition (CVPR)}, pages 447--456, 2015.

\bibitem{he2020momentum}
K. {He}, H. {Fan}, Y. {Wu}, S. {Xie}, and R. {Girshick}.
\newblock Momentum contrast for unsupervised visual representation learning.
\newblock In {\em 2020 IEEE/CVF Conference on Computer Vision and Pattern
  Recognition (CVPR)}, pages 9726--9735, 2020.

\bibitem{he2017mask}
Kaiming He, Georgia Gkioxari, Piotr Dollár, and Ross Girshick.
\newblock Mask r-cnn.
\newblock In {\em 2017 IEEE International Conference on Computer Vision
  (ICCV)}, pages 2980--2988, 2017.

\bibitem{henaff2020data}
Olivier Henaff.
\newblock Data-efficient image recognition with contrastive predictive coding.
\newblock In {\em International Conference on Machine Learning}, pages
  4182--4192. PMLR, 2020.

\bibitem{hinton2002stochastic}
Geoffrey~E Hinton and Sam Roweis.
\newblock Stochastic neighbor embedding.
\newblock In S. Becker, S. Thrun, and K. Obermayer, editors, {\em Advances in
  Neural Information Processing Systems}, volume~15. MIT Press, 2002.

\bibitem{hjelm2019learning}
Devon Hjelm, Alex Fedorov, Samuel Lavoie-Marchildon, Karan Grewal, Philip
  Bachman, Adam Trischler, and Yoshua Bengio.
\newblock Learning deep representations by mutual information estimation and
  maximization.
\newblock In {\em ICLR 2019}. ICLR, April 2019.

\bibitem{hu2021adco}
Qianjiang Hu, Xiao Wang, Wei Hu, and Guo-Jun Qi.
\newblock Adco: Adversarial contrast for efficient learning of unsupervised
  representations from self-trained negative adversaries.
\newblock In {\em 2021 IEEE/CVF Conference on Computer Vision and Pattern
  Recognition (CVPR)}, pages 1074--1083, 2021.

\bibitem{huang2023mast}
Chen Huang, Hanlin Goh, Jiatao Gu, and Joshua~M. Susskind.
\newblock {MAST}: Masked augmentation subspace training for generalizable
  self-supervised priors.
\newblock In {\em The Eleventh International Conference on Learning
  Representations}, 2023.

\bibitem{huang2022learning}
Lang Huang, Shan You, Mingkai Zheng, Fei Wang, Chen Qian, and Toshihiko
  Yamasaki.
\newblock Learning where to learn in cross-view self-supervised learning.
\newblock In {\em 2022 IEEE/CVF Conference on Computer Vision and Pattern
  Recognition (CVPR)}, pages 14431--14440, 2022.

\bibitem{ioffe2015batch}
Sergey Ioffe and Christian Szegedy.
\newblock Batch normalization: Accelerating deep network training by reducing
  internal covariate shift.
\newblock In Francis Bach and David Blei, editors, {\em Proceedings of the 32nd
  International Conference on Machine Learning}, volume~37 of {\em Proceedings
  of Machine Learning Research}, pages 448--456, Lille, France, 07--09 Jul
  2015. PMLR.

\bibitem{jang2023self}
Jiho Jang, Seonhoon Kim, Kiyoon Yoo, Chaerin Kong, Jangho Kim, and Nojun Kwak.
\newblock Self-distilled self-supervised representation learning.
\newblock In {\em 2023 IEEE/CVF Winter Conference on Applications of Computer
  Vision (WACV)}, pages 2828--2838, 2023.

\bibitem{ji2019invariant}
X. {Ji}, A. {Vedaldi}, and J. {Henriques}.
\newblock Invariant information clustering for unsupervised image
  classification and segmentation.
\newblock In {\em 2019 IEEE/CVF International Conference on Computer Vision
  (ICCV)}, pages 9864--9873, 2019.

\bibitem{khosla2020supervised}
Prannay Khosla, Piotr Teterwak, Chen Wang, Aaron Sarna, Yonglong Tian, Phillip
  Isola, Aaron Maschinot, Ce Liu, and Dilip Krishnan.
\newblock Supervised contrastive learning.
\newblock In H. Larochelle, M. Ranzato, R. Hadsell, M.F. Balcan, and H. Lin,
  editors, {\em Advances in Neural Information Processing Systems}, volume~33,
  pages 18661--18673. Curran Associates, Inc., 2020.

\bibitem{kim2021self}
Kyungyul Kim, ByeongMoon Ji, Doyoung Yoon, and Sangheum Hwang.
\newblock Self-knowledge distillation with progressive refinement of targets.
\newblock In {\em 2021 IEEE/CVF International Conference on Computer Vision
  (ICCV)}, pages 6547--6556, 2021.

\bibitem{kim2020mixco}
Sungnyun Kim, Gihun Lee, Sangmin Bae, and Se-Young Yun.
\newblock Mixco: Mix-up contrastive learning for visual representation.
\newblock {\em arXiv preprint arXiv:2010.06300}, 2020.

\bibitem{koohpayegani2021mean}
Soroush~Abbasi Koohpayegani, Ajinkya Tejankar, and Hamed Pirsiavash.
\newblock Mean shift for self-supervised learning.
\newblock In {\em 2021 IEEE/CVF International Conference on Computer Vision
  (ICCV)}, pages 10306--10315, 2021.

\bibitem{Le2015TinyIV}
Ya Le and X. Yang.
\newblock Tiny imagenet visual recognition challenge.
\newblock {\em CS 231N}, 2015.

\bibitem{li2021prototypical}
Junnan Li, Pan Zhou, Caiming Xiong, and Steven Hoi.
\newblock Prototypical contrastive learning of unsupervised representations.
\newblock In {\em International Conference on Learning Representations}, 2021.

\bibitem{li2020addressing}
Tianhong Li, Lijie Fan, Yuan Yuan, Hao He, Yonglong Tian, Rogerio Feris, Piotr
  Indyk, and Dina Katabi.
\newblock Addressing feature suppression in unsupervised visual
  representations.
\newblock {\em arXiv e-prints}, pages arXiv--2012, 2020.

\bibitem{lin2014microsoft}
Tsung-Yi Lin, Michael Maire, Serge Belongie, James Hays, Pietro Perona, Deva
  Ramanan, Piotr Doll{\'a}r, and C.~Lawrence Zitnick.
\newblock Microsoft coco: Common objects in context.
\newblock In David Fleet, Tomas Pajdla, Bernt Schiele, and Tinne Tuytelaars,
  editors, {\em Computer Vision -- ECCV 2014}, pages 740--755, Cham, 2014.
  Springer International Publishing.

\bibitem{lloyd1982least}
S. Lloyd.
\newblock Least squares quantization in pcm.
\newblock {\em IEEE Transactions on Information Theory}, 28(2):129--137, 1982.

\bibitem{navaneet2022constrained}
K.~L. Navaneet, Soroush Abbasi~Koohpayegani, Ajinkya Tejankar, Kossar
  Pourahmadi, Akshayvarun Subramanya, and Hamed Pirsiavash.
\newblock Constrained mean shift using distant yet related neighbors
  for representation learning.
\newblock In Shai Avidan, Gabriel Brostow, Moustapha Ciss{\'e}, Giovanni~Maria
  Farinella, and Tal Hassner, editors, {\em Computer Vision -- ECCV 2022},
  pages 23--41, Cham, 2022. Springer Nature Switzerland.

\bibitem{peng2022balanced}
Xiaokang Peng, Yake Wei, Andong Deng, Dong Wang, and Di Hu.
\newblock Balanced multimodal learning via on-the-fly gradient modulation.
\newblock In {\em 2022 IEEE/CVF Conference on Computer Vision and Pattern
  Recognition (CVPR)}, pages 8228--8237, 2022.

\bibitem{rame2021ixmo}
Alexandre Ramé, Rémy Sun, and Matthieu Cord.
\newblock Mixmo: Mixing multiple inputs for multiple outputs via deep
  subnetworks.
\newblock In {\em 2021 IEEE/CVF International Conference on Computer Vision
  (ICCV)}, pages 803--813, 2021.

\bibitem{ren2015faster}
Shaoqing Ren, Kaiming He, Ross Girshick, and Jian Sun.
\newblock Faster r-cnn: Towards real-time object detection with region proposal
  networks.
\newblock In C. Cortes, N. Lawrence, D. Lee, M. Sugiyama, and R. Garnett,
  editors, {\em Advances in Neural Information Processing Systems}, volume~28.
  Curran Associates, Inc., 2015.

\bibitem{robinson2021contrastive}
Joshua~David Robinson, Ching-Yao Chuang, Suvrit Sra, and Stefanie Jegelka.
\newblock Contrastive learning with hard negative samples.
\newblock In {\em International Conference on Learning Representations}, 2021.

\bibitem{wang2021dense}
Xinlong Wang, Rufeng Zhang, Chunhua Shen, Tao Kong, and Lei Li.
\newblock Dense contrastive learning for self-supervised visual pre-training.
\newblock In {\em Proceedings of the IEEE/CVF Conference on Computer Vision and
  Pattern Recognition}, pages 3024--3033, 2021.

\bibitem{wu2019detectron2}
Yuxin Wu, Alexander Kirillov, Francisco Massa, Wan-Yen Lo, and Ross Girshick.
\newblock Detectron2.
\newblock \url{https://github.com/facebookresearch/detectron2}, 2019.

\bibitem{wu2018unsupervised}
Zhirong Wu, Yuanjun Xiong, Stella~X. Yu, and Dahua Lin.
\newblock Unsupervised feature learning via non-parametric instance
  discrimination.
\newblock In {\em 2018 IEEE/CVF Conference on Computer Vision and Pattern
  Recognition}, pages 3733--3742, 2018.

\bibitem{xie2021propagate}
Zhenda Xie, Yutong Lin, Zheng Zhang, Yue Cao, Stephen Lin, and Han Hu.
\newblock Propagate yourself: Exploring pixel-level consistency for
  unsupervised visual representation learning.
\newblock In {\em Proceedings of the IEEE/CVF Conference on Computer Vision and
  Pattern Recognition}, pages 16684--16693, 2021.

\bibitem{xu2022seed}
Haohang Xu, Xiaopeng Zhang, Hao Li, Lingxi Xie, Wenrui Dai, Hongkai Xiong, and
  Qi Tian.
\newblock Seed the views: Hierarchical semantic alignment for contrastive
  representation learning.
\newblock {\em IEEE Transactions on Pattern Analysis and Machine Intelligence},
  45(3):3753--3767, 2023.

\bibitem{cvpr19unsupervised}
M. {Ye}, X. {Zhang}, P.~C. {Yuen}, and S. {Chang}.
\newblock Unsupervised embedding learning via invariant and spreading instance
  feature.
\newblock In {\em 2019 IEEE/CVF Conference on Computer Vision and Pattern
  Recognition (CVPR)}, pages 6203--6212, 2019.

\bibitem{yoshihashi2022ladder}
Ryota Yoshihashi, Shuhei Nishimura, Dai Yonebayashi, Yuya Otsuka, Tomohiro
  Tanaka, and Takashi Miyazaki.
\newblock Ladder siamese network: a method and insights for multi-level
  self-supervised learning.
\newblock {\em arXiv preprint arXiv:2211.13844}, 2022.

\bibitem{zhang2019your}
Linfeng Zhang, Jiebo Song, Anni Gao, Jingwei Chen, Chenglong Bao, and Kaisheng
  Ma.
\newblock Be your own teacher: Improve the performance of convolutional neural
  networks via self distillation.
\newblock In {\em 2019 IEEE/CVF International Conference on Computer Vision
  (ICCV)}, pages 3712--3721, 2019.

\bibitem{zhao2020makes}
Nanxuan Zhao, Zhirong Wu, Rynson~WH Lau, and Stephen Lin.
\newblock What makes instance discrimination good for transfer learning?
\newblock {\em arXiv preprint arXiv:2006.06606}, 2020.

\bibitem{zheng2021ressl}
Mingkai Zheng, Shan You, Fei Wang, Chen Qian, Changshui Zhang, Xiaogang Wang,
  and Chang Xu.
\newblock Ressl: Relational self-supervised learning with weak augmentation.
\newblock In M. Ranzato, A. Beygelzimer, Y. Dauphin, P.S. Liang, and J.~Wortman
  Vaughan, editors, {\em Advances in Neural Information Processing Systems},
  volume~34, pages 2543--2555. Curran Associates, Inc., 2021.

\bibitem{zheng2021resslv2}
Mingkai Zheng, Shan You, Fei Wang, Chen Qian, Changshui Zhang, Xiaogang Wang,
  and Chang Xu.
\newblock Ressl: Relational self-supervised learning with weak augmentation.
\newblock {\em arXiv preprint arXiv:2107.09282}, 2021.

\end{thebibliography}
}

\end{document}